\documentclass[runningheads]{llncs}
\usepackage[T1]{fontenc}
\usepackage[colorinlistoftodos]{todonotes}
\usepackage{booktabs}
\usepackage{amsmath} 
\usepackage{algorithm}
\usepackage{algorithmicx}
\usepackage{algpseudocode}
\usepackage{amsfonts} 
\usepackage{comment}
\usepackage{multirow}
\usepackage[hidelinks]{hyperref}
\usepackage{graphicx}
\usepackage{tcolorbox}
\usepackage{enumitem}

\begin{document}
\title{The benefits of query-based KGQA systems for complex and temporal questions in LLM era}
%
\titlerunning{Query-based KGQA systems for complex and temporal questions}
%
\author{Artem Alekseev\inst{1}
\and
Mikhail Chaichuk \inst{2,3}
\and
Miron Butko \inst{4}
\and
Alexander Panchenko \inst{1,2,3} 
\and Elena Tutubalina \inst{2,3,5}
\and Oleg Somov \inst{2,6}
\thanks{Corresponding author: \texttt{somov@airi.net}}}

\authorrunning{A. Alekseev et al.}
\institute{Skoltech, Moscow, Russia \and
AIRI, Moscow, Russia \and HSE University, Moscow, Russia \and SaluteDevices, Moscow, Russia \and Kazan Federal University, Kazan, Russia \and
Moscow Institute of Physics and Technology, Moscow, Russia}
\maketitle              
\begin{abstract}
Large language models excel in question-answering (QA) yet still struggle with multi-hop reasoning and temporal questions. Query-based knowledge graph QA (KGQA) offers a modular alternative by generating executable queries instead of direct answers. We explore multi-stage query-based framework for WikiData QA, proposing multi-stage approach that enhances performance on challenging multi-hop and temporal benchmarks. Through generalization and rejection studies, we evaluate robustness across multi-hop and temporal QA datasets. Additionally, we introduce a novel entity linking and predicate matching method using CoT reasoning. Our results demonstrate the potential of query-based multi-stage KGQA framework for improving multi-hop and temporal QA with small language models. 

\textbf{Code and data}: \urlstyle{tt}\url{https://github.com/ar2max/NLDB-KGQA-System}.

\end{abstract}

\section{Introduction}

Modern question-answering (QA) systems based on large language models (LLMs) achieve strong performance on open-domain benchmarks \cite{gao2023retrieval,shinn2023reflexion,touvron2023llama}. These models rely on retrieving information either from internal models parameters or external memory such as vector storage \cite{lewis2020retrieval}. 

Despite their success, QA systems face limitations in both knowledge retrieval mechanisms. Internally, knowledge gaps arise due to uneven fact distribution in pretraining data  -- some facts are more popular in pre-training data than others \cite{allen2023physics}. Externally, retrieval effectiveness is closely tied to document representation, where issues can stem from dense retrieval methods \cite{karpukhin-etal-2020-dense} or suboptimal initial document knowledge representation. For instance, some studies indicate that models struggle to infer inverted knowledge directly from the data \cite{berglund2023reversal}. Thus LLM-based QA systems often struggle with answering rare, multi-hop, or time-sensitive questions -- especially those requiring structured data, despite the possibility of answering such question with internal or external storage \cite{jiang2021can}.

In contrast, parsing-based QA systems generate structured queries that directly retrieve information from structured knowledge bases (KBs), overcoming the limitations of memory-based approaches. Parsing-based systems construct executable queries, ensuring precise and deterministic retrieval from structured databases. While these approaches face challenges related to compositional generalization and domain adaptation \cite{gu2021beyond,somov-tutubalina-2023-shifted,jiang2022knowledge}, they offer key advantages in terms of scalability, accuracy, and efficiency when answering open-domain questions that involve structured data \cite{yu2022decaf,brei2024leveraging}.

In this work, we present a query-based multi-stage pipeline QA system that generates executable queries for retrieving answers from structured knowledge sources. Our system is evaluated across key metrics, including generalization to new domains, rejection ability, and overall system performance on multi-hop and temporal questions. Experimental results demonstrate that our approach achieves frontier performance on selected datasets, surpassing both ChatGPT’s direct answer generation methods and query generation method, which produces executable SPARQL queries.

Our research specifically focuses on Wikidata \cite{vrandevcic2014wikidata} due to its unique characteristics, which present distinct challenges for QA systems. Unlike DBpedia \cite{auer2007dbpedia}, where entities are directly linked to textual literals, Wikidata requires mapping to unique numeric identifiers (Q-IDs), significantly increasing the complexity of Entity Linking. Additionally, Wikidata’s structured representation, including qualifiers and references, introduces higher reasoning complexity in query execution. Moreover, Wikidata is continuously updated, ensuring access to up-to-date information which is crucial factor for real-world applications.

To the best of our knowledge, no prior work has conducted a comprehensive evaluation of a query-based KGQA system on Wikidata in comparison with proprietary LLMs. In this study, we evaluate each component of the multi-stage pipeline independently, as well as assess the end-to-end performance of the entire system — from input question to predicted answer entity IDs. Our findings demonstrate the effectiveness of the proposed pipeline across multiple datasets, highlighting its strong generalization capabilities and improved rejection handling. These results underscore the significance of our approach in advancing structured QA systems and bridging the gap between natural language queries and structured knowledge retrieval.

\section{Related Work}

Knowledge graph question answering (KGQA) has been a longstanding area of research in natural language processing (NLP). However, with the emergence of LLMs, KGQA has gained increased utility. Early approaches, which relied on rule-based systems and traditional semantic parsing components, demonstrated high precision for the input questions but suffered from low recall. In contrast, LLMs are capable of processing large volumes of input data and generating accurate queries, although they are often plagued by generalization issues and hallucinations. Modern KGQA systems based on large language models aim to mitigate these issues by leveraging advanced techniques to improve generalization and reduce hallucinations.

Query-based KGQA systems were initially trained on text-to-SPARQL pairs, incorporating intermediate steps like predicate classification, entity linking, and slot-filling \cite{lysyuk2024konstruktor,mohammed-etal-2018-strong,hu2023empirical}. These approaches focused on simple one-hop questions but remain relevant due to their structured methodology. Efforts to overcome the one-hop limitation include in-context-learning (ICL) with proprietary LLMs \cite{tan2023can} and fine-tuning approaches \cite{banerjee2022modern,brei2024leveraging}.

Research on ICL improvements includes retrieval-augmented generation with few-shot examples \cite{ghajari2024querying}, SPARQL decomposition into atomic sub-steps \cite{li2024framework}, and interactive multi-hop question decomposition \cite{xiong2024interactive}. Fine-tuning strategies involve augmenting text-to-SPARQL pairs \cite{rangel2024sparql}, structured pre-training \cite{qi2024enhancing}, and incorporating linguistic features like Part-of-Speech tags and dependency parsing \cite{muennighoff2202sgpt}.

Ensuring the reliability of SPARQL queries in KGQA involves pre-execution validation to prevent errors. AI-based methods enhance reliability by employing syntactic and semantic query checks \cite{zahera2024sparql}, LLM-driven Chain-of-Thought (CoT) prompting \cite{sun2024kgqa}, error detection \cite{schwabe2025qnl,somov2025}, and graph-based logical consistency verification \cite{baazouzi2024sparql}. Transformer models compare generated queries against reference sets to flag logical flaws \cite{chong2024transkgqa}, while hybrid symbolic-ML methods integrate rule-based reasoning with machine learning for enhanced accuracy \cite{longwell2024tripleglm}. These approaches collectively improve KGQA robustness, reducing incorrect or nonsensical query execution before reaching the user.

As noted in \cite{brei2024leveraging}, proprietary ICL models like ChatGPT-4 or Claude are expensive due to large input sizes and multiple query requests. We build upon this by developing an multi-stage KGQA system using small fine-tuned models, trainable on consumer-grade GPUs. To our knowledge, no prior research has systematically evaluated the generalization and rejection accuracy of such a system, particularly for Wikidata-based temporal and multi-hop QA.

\section{Datasets and Evaluation}

We evaluate our system on four multi-hop and temporal datasets: LC-QuAD 2.0 \cite{dubey2019lc}, RuBQ 2.0 \cite{rybin2021rubq}, QALD-10 \cite{usbeck2024qald}, and PAT-Questions \cite{meem2024pat}. All samples from selected datasets consists of question, SPARQL, mentioned entities and corresponding Q-IDs. The SPARQL can be executed over Wikidata Query engine\footnote{\url{https://query.wikidata.org}}. The key statistics on the complexity for these datasets are presented in Table \ref{tab:dataset_stats}.



\begin{table}[t!]
\centering
\caption{The key statistics for selected benchmarks.}
\label{tab:dataset_stats}
\begin{tabular}{@{}lccccc@{}}
\toprule
\textbf{Dataset} &
  \hspace{2mm}\textbf{Train}\hspace{2mm} &
  \hspace{2mm}\textbf{Test}\hspace{2mm} &
  \hspace{2mm}\textbf{Multi-HOP \%}\hspace{2mm} &
  \hspace{2mm}\textbf{Avg. Hops}\hspace{2mm} &
  \hspace{2mm}\textbf{Hops Range} \\ \midrule
\textbf{QALD-10}       & 402  & 386 & 58.3\% & 2.1 & 1--8  \\
\textbf{LC-QuAD 2.0} & 17,864 & 4,541 & 62.9\% & 1.91 & 1--4      \\
\textbf{RuBQ 2.0}       & 1,920  & 580 & 10\%   & 1.1 & 1--2  \\
\textbf{PAT}        & 4,795 & 1,199 & 53.3\% & 1.68 & 1--2 \\
\bottomrule
\end{tabular}
\end{table}

\textbf{LC-QuAD 2.0} is a benchmark for KGQA over Wikidata, evaluating complex SPARQL-based query generation. It includes multi-hop reasoning, boolean queries, count-based queries, and qualifier-dependent questions. \textbf{QALD-10} is a multilingual benchmark for KGQA, evaluating systems querying Wikidata using SPARQL. We use a subset of English questions for our evaluation. \textbf{RuBQ 2.0} is designed for querying Wikidata with structured SPARQL queries. It includes one-hop, multi-hop, count-based, literal-based, and unanswerable questions.  The dataset covers comparative, temporal, count-based, and superlative queries.  \textbf{PAT} is a benchmark for Present-Anchored Temporal Question Answering, focusing on time-sensitive questions with evolving answers. It includes single-hop and multi-hop questions and features an auto-updating mechanism for answer freshness.

We evaluate our entity linking quality and system's ability to generate accurate SPARQL queries for KGQA by executing predicted and ground truth queries on Wikidata and comparing their entity sets $\hat{\mathcal{T}}$ and $\mathcal{T}$. Accuracy ($Acc$), precision ($p$), recall ($r$), and F1 score are computed based on the intersection of these sets: 
\[
    \mathcal{C} = \mathcal{T} \cap \widehat{\mathcal{T}}, \quad
    p = \frac{|\mathcal{C}|}{|\widehat{\mathcal{T}}|}, \quad
    r = \frac{|\mathcal{C}|}{|\mathcal{T}|}, \quad
    F_1 = \frac{2 p r}{p + r}, \quad
    \text{Acc@1} = 
    \begin{cases}
        1, & \text{if } \mathcal{T} \cap \widehat{\mathcal{T}} = \mathcal{T} \\
        0, & \text{otherwise.}
    \end{cases} \quad
\]
As the key performance measure we use F1 score. It is important to note, that we consider correct execution match if there is full intersection between predicted and gold answer sets at the first SPARQL generation execution. 


\section{KGQA System}

\begin{figure}[t!]
    \centering
    \includegraphics[width=1.0\textwidth]{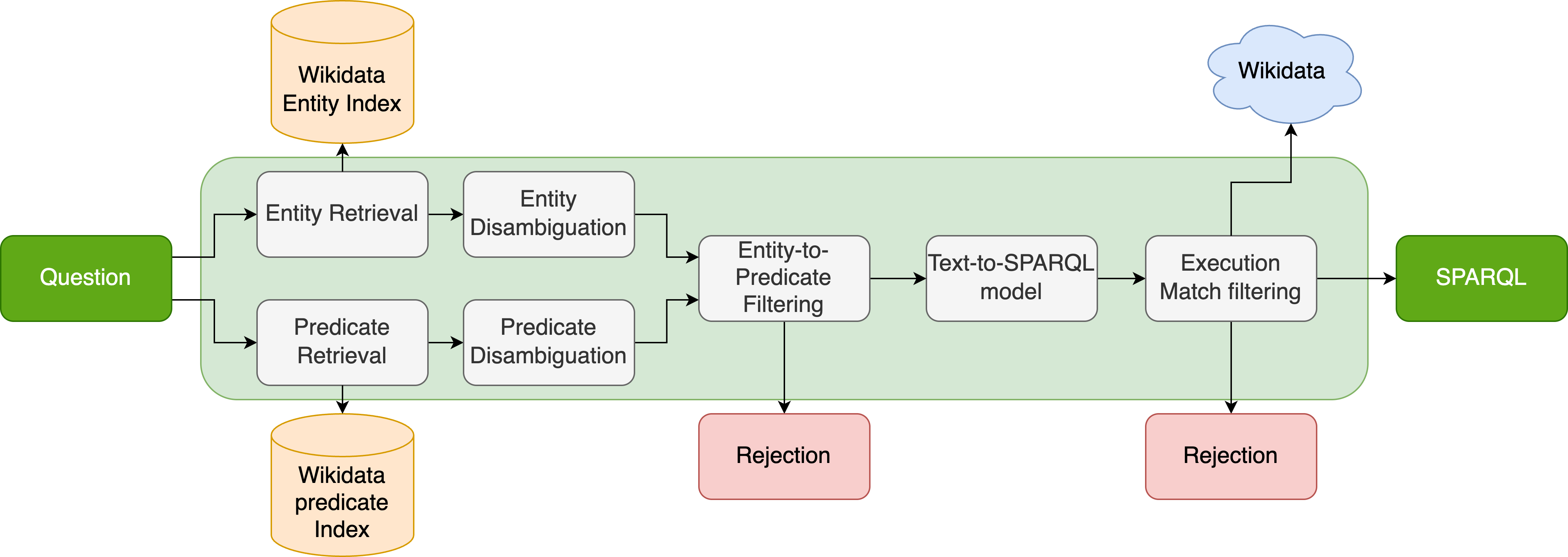} 
    \caption{The proposed KGQA system processes natural language question and translates it into SPARQL for retrieving information from Wikidata. It follows a structured pipeline where entities and predicates are first retrieved and disambiguated using Wikidata indexes. A filtering step then ensures correct entity-predicate mappings before passing it to a text-to-SPARQL model. The generated query undergoes execution match filtering, ensuring valid and accurate results. If retrieval or disambiguation fails, the system rejects the query.}
    \label{fig:kgqa_pipeline}
\end{figure}

We present a query-based KGQA system, shown in Figure \ref{fig:kgqa_pipeline}, designed to generate SPARQL queries for Wikidata. The system includes three core components: (1) entity detection and disambiguation, (2) predicate detection and disambiguation, and (3) an auto-regressive query generation model. We evaluate multiple entity linking approaches -- reasoning-based and fine-tuned solution -- and evaluate their impact on overall system performance. Additionally, we enhance the pipeline through text-to-SPARQL model fine-tuning or few-shot training. Our experiments focus on multi-hop and temporal question-answering, assessing the system’s generalization and ability to reject incorrect queries. The following sections detail each component and its independent and joint metrics.

\subsection{Entity Linking} \label{sec:el}

Entity linking involves identifying the span of an entity mention in a text and linking it to a unique identifier in a knowledge base. In open-domain KGQA systems, such as those over Wikidata, this task is challenging due to the large number of potential entities (over 80 million). We explore two approaches for entity linking: fine-tuned entity linking and entity linking with reasoning models.

\textbf{E2E Entity Linking.} For entity linking, we use the ReFinED model \cite{ayoola-etal-2022-refined}, which links entities using fine-grained entity types and descriptions in a single forward pass. The model performs mention detection, entity typing, and disambiguation, with training in a multi-task manner across four tasks. These loss tasks are combined with weighted loss functions. ReFinED outperforms other models in zero-shot entity linking and has near-perfect coverage of Wikidata for our datasets. We applied it without modification, using the original code repository for inference\footnote{\url{https://github.com/amazon-science/ReFinED}}. For the comparison we also evaluate Falcon \cite{falcon_el} and Spacy\footnote{\url{https://spacy.io/api/entitylinker}} open-source entity linking tools.


\textbf{Entity Linking with Reasoning Models.} While ReFinED provides a strong baseline, it struggles with less popular entities and dataset generalization. To address these, we propose a two-step approach \textbf{RetReason}: \textbf{entity retrieval} and \textbf{entity disambiguation} method.

\textbf{Entity Retrieval.} We construct a BM25 index over Wikidata, filtering out entities with fewer than ten predicates resulting in 15 million entity index size (we do that to preserve only popular entities and keep the index in reasonable size). We optimize BM25 hyperparameters ($k_1$, $b$) for each dataset, favoring recall. We retrieve the top 10 and 100 candidate entities. The retriever hyperparams and quality is in Appendix \ref{app:B}.

\textbf{Entity Disambiguation.} Using a Chain-of-Thought reasoning approach, we disambiguate retrieved entities by prompting reasoning models (DeepSeek-R1 and ChatGPT-4) to select the correct entity. Table \ref{tab:entity_linking} presents F1 scores for all models. The results demonstrate that Refined entity linking approaches surpasses other methods, but RetReason is competitive and can be utilized through LLM API. The prompt for entity disambiguation for CoT reasoning is provided in the Appendix \ref{app:A}.

\begin{table}[t!]
\centering
\caption{Entity Linking F1 performance for all methods.}
\label{tab:entity_linking}
\begin{tabular}{@{}lcccc@{}}
\toprule
\textbf{Method} & \textbf{QALD-10} & \textbf{LC-QuAD 2.0} & \textbf{RuBQ 2.0} & \textbf{PAT} \\ \midrule
Refined & \textbf{65.0} & \textbf{57.4} & \textit{57.2} & \textbf{93.1} \\
Spacy & 49.0 & 46.0 & 31.0 & 40.0 \\
Falcon & 33.0 & 43.0 & 33.0 & 35.0 \\\midrule
\begin{tabular}[c]{@{}l@{}}RetReason with DeepSeek@10\end{tabular} & 47.43 & 47.87 & 49.44 & 39.83 \\
\begin{tabular}[c]{@{}l@{}}RetReason with ChatGPT@10\end{tabular} & 50.57 & 48.65 & 52.58 & 43.82 \\\midrule
\begin{tabular}[c]{@{}l@{}}RetReason with DeepSeek@100\end{tabular} & 51.61 & 50.64 & 54.97 & 40.22 \\
\begin{tabular}[c]{@{}l@{}}RetReason with ChatGPT@100\end{tabular} & \textit{53.86} & \textit{54.14} & \textbf{60.15} & \textit{44.99} \\\bottomrule
\end{tabular}
\begin{center}
\caption{Predicate F1 Matching for RetReason.}
\label{tab:predicate_matching}
\begin{tabular}{@{}lcccc@{}}
\toprule
\textbf{Method} & \textbf{QALD-10} & \textbf{LC-QuAD 2.0} & \textbf{RuBQ 2.0} & \textbf{PAT} \\ \midrule
\begin{tabular}[c]{@{}l@{}}RetReason with DeepSeek@10\end{tabular} & 55.21 & \textbf{65.81} & 67.86 & \textbf{43.49} \\
\begin{tabular}[c]{@{}l@{}}RetReason with ChatGPT@10\end{tabular} & 58.75 & 63.43 & \textbf{73.13} & 41.24 \\ \midrule
\begin{tabular}[c]{@{}l@{}}RetReason with DeepSeek@100\end{tabular} & 56.27 & 62.97 & 64.35 & 39.29 \\
\begin{tabular}[c]{@{}l@{}}RetReason with ChatGPT@100\end{tabular} & \textbf{62.89} & 62.95 & \textbf{72.83} & 35.43 \\ \bottomrule
\end{tabular}
\end{center}
\end{table}

\subsection{Predicate Matching}

We use the entity retrieval and disambiguation methodology for predicate selection. 
While classification-based baselines are effective for one-hop questions, they cannot be applied to multi-hop cases because the number of relations is not fixed -- we need to do arbitrary predicate selection among all WikiData predicates. 

\textbf{Predicate Retrieval.} As in the Entity Retrieval section, we follow the same matching methodology. We construct a BM25 search index over all Wikidata predicates, resulting in over 12k predicates index size. Each predicate is represented by its label and description texts. The search procedure is identical to Entity Retrieval. We retrieve the top 10 and 100 candidate predicates and evaluate retrieval effectiveness using recall as the primary metric. The hyper-parameter search and search quality are presented in Appendix \ref{app:B}.

\textbf{Predicate Disambiguation.} Similar to Entity Disambiguation, we address the predicate selection problem by identifying the relevant predicates from the set of retrieved candidates that are most related to the ones mentioned in the question. The predicate disambiguation step is performed similarly as entity disambiguation, except that the system prompt and corresponding descriptions of each predicate, coming from the retrieval model, are used. The prompt for entity disambiguation for CoT reasoning is provided in the Appendix \ref{app:A}. The final metrics for predicate disambiguation presented in Table \ref{tab:predicate_matching}.

\subsection{Text-to-SPARQL Modeling}\label{sec:t2s}

\begin{figure}[t!]
    \centering
    \includegraphics[width=1.0\textwidth]{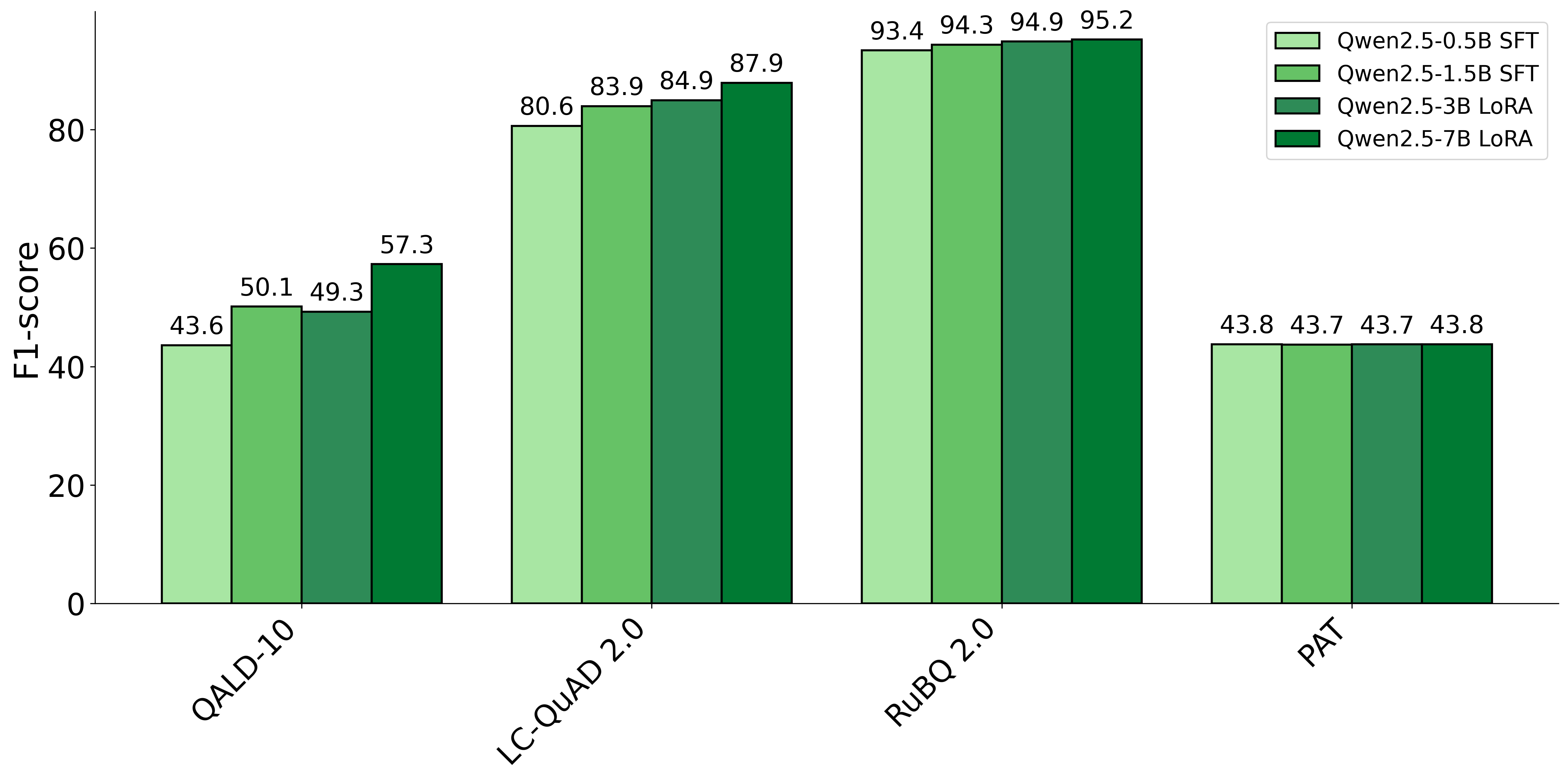} 
    \caption{The models comparison across different sizes (0.5B, 1.5B, 3B, 7B) of Qwen2.5 models and training approaches (LoRA \& SFT) with F1 score.}
    \label{fig:model_comparison}
\end{figure}

For the prediction of the final Text-to-SPARQL model, we apply fine-tuning of pretrained LLM's from Qwen2.5\footnote{\url{https://huggingface.co/collections/Qwen/qwen25-66e81a666513e518adb90d9e}} family for moderate sizes -- 0.5B, 1.5B, 3B and 7B to evaluate the generalization performance on different scales. We do full-supervised fine-tuning for 0.5B and 1.5B models and LoRA for the 3B and 7B. The model is trained to predict the final SPARQL query, given small query generation instruction, question, candidate entity set from entity linking stage and predicate entity set from predicate linking stage. The training results for selected datasets and for models of different sizes are present in Figure \ref{fig:model_comparison}. The figure does not exhibit a consistent trend of quality improvement with an increase in scale across all datasets.
During training stage, we want to emulate the inference phase, where entity linking usually gives more entities, then expected. We combine incorrect entities with the correct ones, based on their BM25 similarity to the gold entity mentioned in the question. The prompt format and the model approach training hyper-parameters are presented in Appendix \ref{app:E}.

\subsection{Question and Predicate Filtering} 

The text-to-SPARQL model is trained on pairs of questions and SPARQL queries, so it will always generate a query for any entity and predicate, even in cases where doing so is fundamentally impossible according to the knowledge graph ontology. This results in a high false positive rate, where incorrect queries are produced, misleading users and requiring manual validation. To overcome this issue, we propose a simple filtering algorithm to filter out Entities and Predicates that do not relate to each other. The algorithm pseudo-code is provided in Appendix \ref{app:F}.

As a second reliability check, we execute the generated SPARQL query and analyze its output. If the query results in an error or returns an empty set, it is considered incorrect. Errors indicate structural or syntax issues, while an empty set suggests that the query does not retrieve any relevant information from the knowledge graph. In both cases, we discard the query and inform the user that an accurate SPARQL query cannot be generated, preventing misleading or unverified results.

\section{Experiments}

The system is evaluated using three approaches: GPT-4o as a direct question-answering (QA) model, GPT-4o as a text-to-SPARQL generation model and our developed system. In our following set of experiments we want to show the advantage of small models for KGQA problem.

\textbf{GPT-4o as a Direct QA Model.} In this approach, GPT-4o is prompted as an open-domain QA system, leveraging few-shot learning to infer factual answers. No constraints are placed on the usage of knowledge, meaning GPT-4o can access any information available through its API. The generated answers are evaluated based on the exact match between gold labels and predicted responses. 

\textbf{Few-shot GPT-4o for SPARQL Query Generation.} In this setup, GPT-4o receives a prompt instructing it to transform natural language questions into valid SPARQL queries. The prompt includes predicted entity, provided by ReFined, and predicate matching, provided by RetReason from previous stage. The generated SPARQL queries are then executed against Wikidata, and their results are compared to the ground truth answers.


\begin{table*}[t!]
    \centering
    \caption{End-to-end comparison of GPT-4o, existing solutions, and the full KGQA system on SPARQL query generation, evaluated by accuracy and F1 score of predicted answer entity IDs given the input question.}
    \label{tab:system_performance}
    \resizebox{\linewidth}{!}{ 
    \begin{tabular}{@{}lcccccccc@{}}
        \toprule
        \textbf{Model} & \multicolumn{2}{c}{\textbf{QALD-10}} & \multicolumn{2}{c}{\textbf{LC-QuAD 2.0}} & \multicolumn{2}{c}{\textbf{RuBQ 2.0}} & \multicolumn{2}{c}{\textbf{PAT}} \\
        \cmidrule(lr){2-3} \cmidrule(lr){4-5} \cmidrule(lr){6-7} \cmidrule(lr){8-9}
        & \textbf{F1} & \textbf{Acc@1} & \textbf{F1} & \textbf{Acc@1} & \textbf{F1} & \textbf{Acc@1} & \textbf{F1} & \textbf{Acc@1} \\
        \midrule
        GPT-4o (Direct QA)   & 37.3 & -  & 21.1 & -  & \textbf{77.0} & -  & 18.3 & -  \\
        \midrule
        Few-shot GPT-4o + ReFineD &  33.8   & 24.1  & 24.4  & 14.9  & 38.3   & 27.2  &  17.7   & 0.0  \\
        \midrule
        SPINACH \cite{liu2024spinach}     &  \textbf{69.5}  & -  &  -   & -  &  -   & -  &  -   & -  \\
        SPARQL-QA \cite{borroto2022sparql}  &  \textit{45.38}  & -  &  -   & -  &  -   & -  &  -   & -  \\
        Konstruktor\cite{lysyuk2024konstruktor}          &  -   & -  &  -   & 21.9  &  -   & 34.4  &  -   & -  \\
        Text2Graph(MEKER Wiki4M)\cite{chekalina2022meker}          &  -   & -  &  -   & 26.6  &  -   & -  &  -   & -  \\
        Review-Then-Refine \cite{chen2024then}   &  -   & -  &  -   & -  &  -   & -  & 33.3  & -  \\
        \midrule
        Qwen-0.5B + RetReason & 22.1 & 21.3  & 22.4 & 21.6  & 38.6 & 37.3  & 28.2 & 0.2  \\
        Qwen-0.5B + ReFineD   & 25.6 & 24.7  & 28.6 & 27.4  & 43.9 & 43.3  & \textit{38.9} & 0.2  \\
        \midrule
        Qwen-1.5B + RetReason & 24.0 & 23.1  & 23.6 & 22.8  & 36.2 & 35.1  & 27.9 & 0.2  \\
        Qwen-1.5B + ReFineD   & 27.6 & \textit{26.8}  & 30.3 & 28.9  & 43.2 & 42.6  & 38.9 & 0.2  \\
        \midrule
        Qwen-3B + RetReason   & 24.1 & 23.1  & 23.7 & 22.8  & 40.4 & 39.1  & 27.8 & 0.2  \\
        Qwen-3B + ReFineD     & 27.6 & 26.3  & \textbf{31.2} & \textbf{29.5}  & 45.2 & \textit{44.8}  & 38.9 & 0.2  \\
        \midrule
        Qwen-7B + RetReason   & 24.9 & 23.8  & 24.7 & 23.8  & 42.5 & 41.1  & 28.1 & 0.2  \\
        Qwen-7B + ReFineD     & 31.0 & \textbf{29.7}  & \textit{31.0} & \textit{29.4}  & \textit{46.8} & \textbf{45.9}  & \textbf{39.0} & 0.2  \\
        \bottomrule
    \end{tabular}
    }
\end{table*}

Table \ref{tab:system_performance} shows that in datasets where simple one-hop (QALD-10 and RuBQ 2.0) questions dominate, direct question answering with ChatGPT yields promising results. However, in datasets primarily composed of temporal and multi-hop questions (LC-QuAD 2.0 and PAT), the pipeline-based KGQA system significantly outperforms both ChatGPT’s direct question answering and established baselines.

\section{Generalization Study} 

To address potential overfitting on the training dataset, we carefully consider the risk that the model may achieve high performance on the test split due to learned dataset-specific patterns rather than genuine generalization. Such patterns naturally arise because each dataset follows a distinct data collection pipeline, resulting in similarities in question structure, domain focus, or other inherent characteristics. To mitigate this issue and ensure robust generalization, we adopt a cross-dataset validation strategy, where the model is trained on three out of four datasets and evaluated on the test split of the fourth one that remains entirely unseen during training. This approach enables a more rigorous assessment of the model’s ability to generalize to novel question distributions while reducing the likelihood of dataset-specific memorization.

\begin{figure}[t!]
    \centering
    \includegraphics[width=1.0\textwidth]{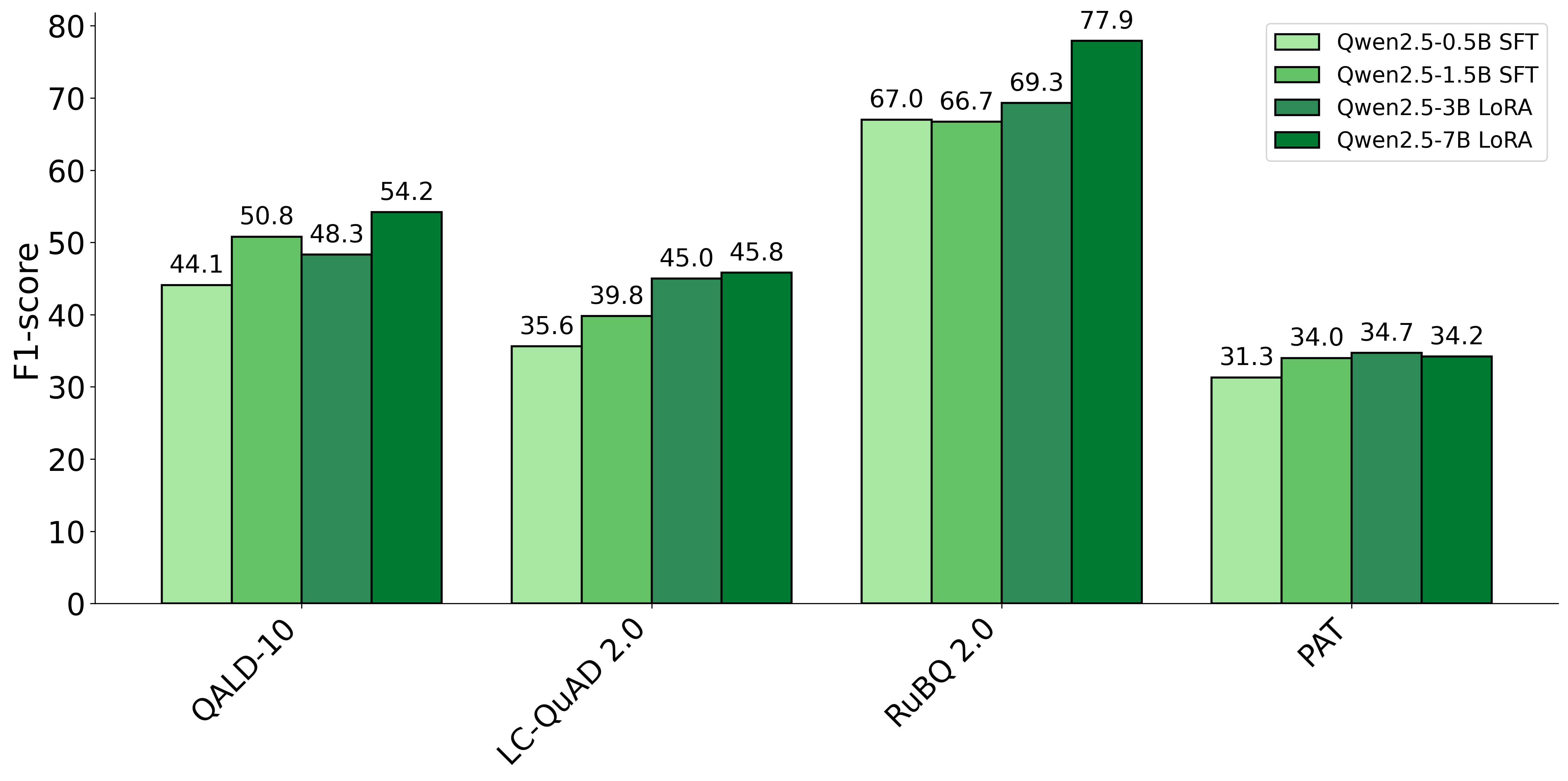} 
    \caption{The generalization evaluation across different sizes (0.5B, 1.5B, 3B, 7B) of Qwen2.5 models and training approaches (LoRA \& SFT) with F1 score.}
    \label{fig:generalization_study}
\end{figure}

The core component of our approach is a text-to-SPARQL model, fine-tuned on the training split of each dataset. In Figure \ref{fig:generalization_study}, we evaluate its generalization capabilities for query generation on an out-of-distribution (OOD) dataset, assuming correct entity linking and predicate matching. The model is trained on three datasets and evaluated on the remaining one, following the same training procedure outlined previously in Section \ref{sec:t2s}. While the models do not surpass the performance of the fine-tuning approach, the performance gap remains relatively small. The largest gap observed on the LC-QuAD 2.0 dataset, indicating the presence of dataset-specific differences in target query structures and the fact that LC-QuAD 2.0 is the largest dataset across all selected.

Furthermore, Table \ref{tab:generalization_end2end} presents an evaluation of the full system in a cross-validation setting across our datasets. For this evaluation, we employed Qwen2.5 (1.5B) along with the ReFineD entity linker. We observe that, in certain cases (QALD-10 and LC-QuAD 2.0), the model exhibits comparable generalization to new datasets, achieving performance levels close to those obtained when fine-tuned on each dataset individually.

\setlength{\tabcolsep}{5pt}
\begin{table}[t!]
    \centering
    \caption{End-to-end generalization performance of the system trained with Qwen2.5-1.5B and ReFineD entity linking, evaluated from question input to predicted answer entity IDs.}
    \begin{tabular}{@{}llccc@{}}
        \toprule
        \textbf{Train} & 
        \textbf{Test} & \textbf{F1} \\
        \midrule
        RuBQ 2.0 + PAT + LC-QuAD 2.0 & QALD-10 & 29.4 \\
        RuBQ 2.0 + QALD-10 + PAT & LC-QuAD 2.0 & 20.8 \\
        QALD-10 + PAT + LC-QuAD 2.0 & RuBQ 2.0 & 22.5 \\
        RuBQ 2.0 + QALD-10 + LC-QuAD 2.0 & PAT & 12.5  \\
        \bottomrule
    \end{tabular}
    \label{tab:generalization_end2end}
\end{table}

\section{Rejection Study} 

Ensuring the reliability of SPARQL query generation is essential for practical KGQA systems. Many models, including GPT-4o, generate responses regardless of their validity, often leading to a high proportion of incorrect answers. To address this issue, we introduce a rejection mechanism designed to identify and filter out erroneous query generations, thereby enhancing system reliability. This study evaluates the effectiveness of our approach in rejecting invalid SPARQL queries and compares it with GPT-4o, which lacks a structured rejection mechanism. Our findings underscore the importance of incorporating early termination and execution-based filtering to mitigate the risk of generating misleading outputs.


\begin{table}[t!]
    \centering
    \caption{The detection of incorrect query generations using rejection mechanisms. Each value represents the proportion of identified incorrect SPARQL queries.}
    \label{tab:rejection}
    \begin{tabular}{@{}lccc@{}}
        \toprule
        \textbf{Dataset}\hspace{2mm} & 
        \hspace{2mm}\textbf{LLM rejection}\hspace{2mm} & 
        \hspace{2mm}\textbf{Execution} \hspace{2mm} & \hspace{2mm}\textbf{Filtering \& Execution}\hspace{2mm} \\
        \midrule
        QALD-10 & 11.42\% & 27.3 \% & 66.3 \% \\
        LC-QuAD 2.0 & 29.7 \% & 51.3 \%  & 72.6 \% \\
        RuBQ 2.0 & 0.0 \% & 41.9 \% & 84.5 \% \\
        PAT & 21.9\% & 40.1 \% & 61.3 \% \\
        \bottomrule
    \end{tabular}
    \label{tab:rejection_table}
\end{table}

We evaluate the system's rejection rate across all datasets and compare the results to direct question answering with LLM rejection. Table \ref{tab:rejection_table} presents the percentage of incorrect SPARQL queries identified using three rejection policies. For direct question answering with ChatGPT, we leverage the model’s internal mechanism to recognize when it cannot answer a question through prompting. For SPARQL generation, we execute the generated queries against the database and filter out those that return no results. The final column reports a joint rejection approach that combines entity-to-predicate filtering with execution-based filtering. As the results indicate, entity-to-predicate filtering (presented in Appendix \ref{app:F}) significantly improves the detection of incorrect query generations by leveraging the constraints of the knowledge graph ontology.



\section{Conclusion}

In this study, we evaluated the performance of the multi-stage KGQA system. As part of the system, we proposed an entity linking approach leveraging the reasoning abilities of LLMs, which performs comparably to traditional entity detection and disambiguation methods. Additionally, we introduced a rejection algorithm that detects errors prior to SPARQL generation. While the system's individual components perform good overall, a common issue observed in systems with successive components is the propagation of errors, which ultimately impacts the overall quality. Despite this, our approach demonstrates a significant improvement over ChatGPT-based QA systems and existing approaches, particularly in complex multi-hop and temporal question tasks, such as those found in the LC-Quad and PAT datasets, underscoring the effectiveness of the proposed method.

\section*{Acknowledgments}

This research was supported in part through computational resources of HPC facilities at HSE University \cite{hse-cluster-link}. The work has been supported by the Russian Science Foundation grant \# 23-11-00358. 

%
%
\bibliographystyle{splncs04}
\bibliography{custom}

\appendix


\section{Entity Disambiguation and Predicate Matching with Reasoning}\label{app:A}

The prompt and evaluation procedure is specified in the solution repository folder ``disambiguation'': \url{https://github.com/ar2max/NLDB-KGQA-System}.

\section{Entity and Predicate Retrieval Hyperparameters}\label{app:B}

\begin{table}[H]
    \centering
    \caption{BM25 hyperparameters (k1 and b) and Recall@10/100 for entity and predicate candidate retrieval. Each cell reports values in the format: entity / predicate.}

    \scalebox{0.80}{
    \begin{tabular}{@{}lcccc@{}}
        \toprule
        \textbf{Dataset} & \textbf{k1} & \textbf{b} & \textbf{Recall@10} & \textbf{Recall@100} \\
        \midrule
        \textbf{QALD-10}       & 2.95 / 5.18 & 0.2 / 0.01 & 0.52 / 0.74 & 0.66 / 0.91 \\
        \textbf{LC-QuAD 2.0}   & 2.45 / 2.95 & 0.2 / 0.01 & 0.56 / 0.89 & 0.71 / 0.98 \\
        \textbf{RuBQ 2.0}      & 1.39 / 2.0 & 0.4 / 0.01 & 0.58 / 0.85  & 0.72 / 0.97 \\
        \textbf{PAT}           & 1.0 / 0.1   & 0.7 / 0.01 & 0.79 / 1.0  & 0.88 / 1.0 \\
        \bottomrule
    \end{tabular}
    }
    \label{tab:bm25_entity_predicate_combined}
\end{table}

\section{Text-to-SPARQL Hyperparameters}\label{app:E}

In an end-to-end setting, entities and relations are extracted with high recall, which often results in the inclusion of additional irrelevant entities and relations alongside the correct ones. To help the model distinguish relevant entities and relations from irrelevant ones, we introduce additional entities and relations during fine-tuning. Specifically, we retrieve the top five most similar entities and relations using BM25 and randomly add them to the gold-standard ones before passing them to the model. This approach encourages the model to disambiguate input information, filtering out unrelated entities while focusing on generating the correct SPARQL structure.
The hyperparameters for the Text-to-SPARQL model training are as follows: the tokenizer max length is set to 1024, with 3 epochs and a batch size of 8. The model uses 500 warmup steps, a weight decay of 0.01, and a learning rate of \(2 \times 10^{-5}\). Gradient accumulation steps are set to 1, and the random seed is 42.


\section{Filtering Algorithm}\label{app:F}

The entity-to-predicate filtering algorithm that rejects the generation based on knowledge graph ontology.

\vspace{-0.5cm}

\begin{algorithm}[H]
\caption{Entity-to-Predicate Filtering Approach}
\label{alg:entity_predicate_filtering}

\begin{algorithmic}[1]

\Require Set of entities $\mathcal{E}$, Set of relations $\mathcal{R}$
\Ensure Boolean flag indicating entity-predicate compatibility

\Function{GetEntityRelations}{$e$}
    \State $\mathcal{R}_{in} \gets \emptyset$, $\mathcal{R}_{out} \gets \emptyset$
    \For{each triple $(s, r, o)$ in the knowledge graph}
        \If{$o = e$} 
            \State $\mathcal{R}_{in} \gets \mathcal{R}_{in} \cup \{r\}$
        \ElsIf{$s = e$}
            \State $\mathcal{R}_{out} \gets \mathcal{R}_{out} \cup \{r\}$
        \EndIf
    \EndFor
    \State \Return $(\mathcal{R}_{in}, \mathcal{R}_{out})$
\EndFunction

\Function{CheckEntityMismatch}{$\mathcal{E}, \mathcal{R}$}
    \State $\text{mismatch} \gets \text{True}$
    \For{each entity $e \in \mathcal{E}$}
        \State $(\mathcal{R}_{in}, \mathcal{R}_{out}) \gets \Call{GetEntityRelations}{e}$
        \If{$(\mathcal{R}_{in} \cup \mathcal{R}_{out}) \cap \mathcal{R} \neq \emptyset$}
            \State $\text{mismatch} \gets \text{False}$
            \State \textbf{break}
        \EndIf
    \EndFor
    \State \Return $\text{mismatch}$
\EndFunction
\end{algorithmic}

\end{algorithm}

\end{document}